%% file: root.tex
\documentclass[letterpaper, 10 pt, conference]{ieeeconf}  

\IEEEoverridecommandlockouts                              

\usepackage{verbatim}
\usepackage{color}
\usepackage{xcolor}
\usepackage{graphicx}
\usepackage{hyperref}
\usepackage{booktabs}
\usepackage{multirow}
\usepackage{listings}
\usepackage{makecell}
\usepackage[normalem]{ulem}
\usepackage{amssymb}
\usepackage{subcaption}
\usepackage{float}
\usepackage{algorithm,algpseudocode}
\usepackage[font=small]{caption}
\usepackage{paralist}
\definecolor{light-gray}{gray}{0.95}

\newcommand{\mypara}[1]{{\smallskip\noindent \bf #1}\hspace{0.1in}}
\title{\LARGE \bf
Robofleet: Open Source Communication and Management for\\ Fleets of Autonomous Robots
}

\author{Kavan Singh Sikand$^{1}$ \and Logan Zartman$^{1}$ \and Sadegh
Rabiee$^{1}$ \and Joydeep Biswas$^{1}$%
\thanks{\noindent$^{1}$Computer Science Department,
        University of Texas at Austin, USA.
        {\tt\small \{kvsikand, logan.zartman, srabiee, joydeepb\} @utexas.edu}}}

\begin{document}

\maketitle
\thispagestyle{empty}
\pagestyle{empty}

\begin{abstract}
\input{abstract}
\end{abstract}

\setlength{\textfloatsep}{1em}
\setlength{\abovecaptionskip}{0.5em}

\begin{figure*}[t]
  \centering
  \includegraphics[width=\textwidth]{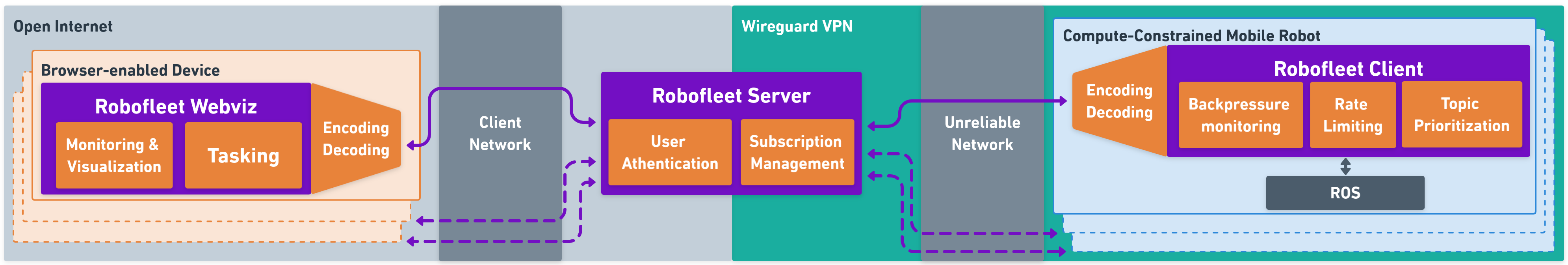}
  \caption{Robofleet System Diagram. The central Robofleet Server facilitates communication between compute-constrained mobile robots on a potentially unreliable network, as well as communication with human robot administrators over the open Internet. }
  \label{fig:system_overview}
\end{figure*}

\input{introduction_short}
\input{related_work_short}

\input{systemoverview}

\input{results}

\input{future_work}

\input{conclusion}

\input{acknowledgements.tex}
\bibliographystyle{IEEEtran}
\bibliography{bibliography}

\end{document}

%% file: abstract.tex
Long-term deployment of a fleet of mobile robots requires
reliable and secure two-way communication channels between individual robots and
remote human operators for supervision and tasking.
Existing open-source solutions to this problem degrade in performance in challenging real-world situations such as intermittent and low-bandwidth connectivity, do not provide security control options, and can be computationally expensive on hardware-constrained mobile robot platforms.        
In this paper, we present Robofleet, a lightweight open-source system which provides inter-robot communication, remote monitoring, and remote tasking for a heterogenous fleet of ROS-enabled service-mobile robots that is designed with the practical goals of resilience to network variance and security control in mind.      	

Robofleet supports multi-user, multi-robot communication via a central server. This architecture deduplicates network traffic between robots, significantly reducing overall network load when compared with native ROS communication. This server also functions as a single entrypoint into the system, enabling security control and user authentication. Individual robots run the lightweight Robofleet client, which is responsible for exchanging messages with the Robofleet server. It automatically adapts to adverse network conditions through backpressure monitoring as well as topic-level priority control, ensuring that safety-critical messages are successfully transmitted. Finally, the system includes a web-based visualization tool that can be run on any internet-connected, browser-enabled device to monitor and control the fleet.

We compare Robofleet to existing methods of robotic communication, and demonstrate that it provides superior resilience to network variance while maintaining performance that exceeds that of widely-used systems.

%% file: introduction_short.tex
\section{INTRODUCTION}

Remote management, multi-agent communication, and user tasking for service-mobile robots is essential for
long-term deployments -- some long-term projects such as the
CoBots~\cite{biswas20161000km} and STRANDS~\cite{hawes2017strands} have relied
on custom remote monitoring and tasking interfaces to fulfil this need, but a
more general open-source solution for arbitrary  and heterogenous fleets of
robots remains elusive.
In this paper, we present Robofleet -- a simple, robust, and reusable solution
to this problem.

An effective multi-robot, multi-user fleet management system must satisfy several key  criteria -- the system must
\begin{inparaenum}
  \item support multiple simultaneously deployed robots,
  \item support communication both between robots as well as operators and robots,
  \item have minimal compute overhead and be capable of running on low-powered devices,
  \item support secure communications and secure access controls, and
  \item be resilient to fluctuating network bandwidth and availability.
\end{inparaenum}

While there is no single open-source solution that meets all such criteria,
several partial solutions for remote robot monitoring and multi-robot
communication include the Robot Web Tools~\cite{toris2015robot}, 
Rosbridge~\cite{Crick2011RosbridgeRF}, and the native
inter-process communication of Robot Operating System (ROS). While
these solutions are  effective at meeting the use case of single-robot
deployments or short-range remote monitoring over a single reliable network,
they exhibit degraded performance in challenging conditions such as intermittent
network connectivity or with a large number of clients.

Robofleet includes several features to meet the aforementioned needs of reliable
multi-robot, multi-user fleet management. It supports message deduplication and
automatic detection of adverse network conditions using backpressure monitoring.
In addition, it supports configuration for rate limiting of topics
combined with priority-based topic scheduling, ensuring that safety-critical
messages take precedence over others. Its single-server architecture prevents
duplication of message streams between robots, further decreasing network load
in the case of multi-robot communication.
Robofleet uses a compact message format to minimize bandwidth usage and enable
high throughput rates when compared with Rosbridge. Robofleet also provides
topic-level access control, user authentication, and supports static IP
address-based traffic control by leveraging a secure VPN. In addition to a
central server, transport layer, and robot client, Robofleet includes an
extensible web-based visualizer and tasking tool tailored towards autonomous
mobile robot deployment, which enables connection to the Robofleet system from
any browser-enabled device. 

We provide experimental results to demonstrate Robofleet's superior performance
compared to existing state-of-the-art solutions in the case of adverse network
conditions and multi-robot interactions. We observe that Robofleet gracefully
recovers from intermittent connectivity $\sim 5$ times faster than
Rosbridge, and is able to maintain near constant latency as the number of robots
increases compared to compared to linear degradation using ROS. Robofleet is
available as open source code at
\url{https://github.com/ut-amrl/robofleet}.

%% file: related_work_short.tex
\section{Related Work}

Beyond the explicit goals of monitoring and tasking, successfully sharing
information between robots enables a wide variety of research initiatives beyond
long-term autonomous deployment. Waibel et al.~\cite{waibel2011roboearth}
introduce a platform consisting of communication
layers~\cite{mohanarajah2014rapyuta} and databases to construct a shared ``world
model" between robots, allowing them to succeed at a wide range of tasks. While
these works seek to allow robots to share information over long time scales, in
this paper we focus on the short-term, time-sensitive information exchange
necessary to enable tasking and monitoring. In addition to long-term autonomous
deployment, multiagent communication is instrumental for applications such
as collaborative mapping~\cite{CoSLAM, CloudMapping}, distributed
control~\cite{wei2021onevision}, and cooperative team behaviors such as robot soccer~\cite{robocup97}.

There has been a significant amount of interest in this short-term communication task in the robotics community, resulting in a variety of widely-used utilities. Robot Web Tools includes an ecosystem of different tools that use the common Rosbridge transport layer. These tools include visualization layers such as Ros2DJS, client libraries such as RosLibJS and RosLibPy, and interactive dashboards \cite{toris2015robot} \cite{Crick2011RosbridgeRF}. There has also been recent research interest in developing interfaces for human operators to interact with fleets of remote robots \cite{Roldn2019MultirobotSV}. Each of these tools individually address discrete parts of the technical pipeline required to successfully deploy autonomous robots, but stitching them together into a coherent workflow currently involves significant overhead. Additionally, the core Rosbridge transport layer has critical performance and bandwidth consumption issues that limit practical use cases of these tools. A common alternative to Rosbridge is using ROS itself to facilitate communication. A shared ROS master can enable fast robot-to-robot communication when compared with Rosbridge. There has been research in systems which enable communication using a shared ROS master between robots across the internet using port forwarding \cite{RosPortForwarding2017}. However, this architecture lacks authentication support, degrades in performance as the number of robots increases, requires that all the machines in the system run ROS, and requires special network configuration.

Robofleet outperforms Rosbridge in adverse network conditions and surpasses the remote ROS network in large fleets of robots in addition to providing features such as topic-level access control and user authentication.

%% file: systemoverview.tex
\section{System Overview}

In this section we present the details of the Robofleet system. Integral to Robofleet is the wire \textbf{serialization} format used for message transmission, which is the universal language that allows the various components of the Robofleet system to communicate with each other. As illustrated by figure \ref{fig:system_overview}, Robofleet consists of three major components.
The \textbf{Robofleet Client} is deployed on each robot in the system, and is responsible for communication between the robot's local ROS system and the Robofleet Server, including handling of adverse network conditions.
The \textbf{Robofleet Server} is responsible for authenticating requests, relaying messages, and maintaining subscription information.
\textbf{Robofleet Webviz} is a browser-based front-end that supports any browser-enabled device and enables robot monitoring and tasking.

\subsection{Serialization}
After a survey of various encoding methods, Robofleet was built using Flatbuffers to encode ROS messages \cite{FlatbuffersWhitePaper}.
This widely-supported binary serialization format is significantly more compact than higher-level representations like JSON, while being more widely accessible than the native ROS binary format. Table \ref{tab:encoding} shows a comparison of these encoding formats for common ROS message types. Flatbuffers supports a variety of languages including C++ and JavaScript, and have guarantees about the backwards compatibility of message binaries. Another common encoding format, protobuf, was also considered, but Flatbuffers has been shown to have significantly faster encoding and decoding performance~\cite{FlatbuffersBenchmarks}, which is critical to our use case.
Robofleet contains a central, extensible serialization library defining encoding and decoding specifications for various ROS message types, which is implemented in both C++ and TypeScript.

\begin{table}
  \caption{\textsc{Comparison of Encoding Sizes for ROS messages}}
  \label{tab:encoding}
  \centering
  \begin{tabular}{lrrr}
      \toprule
      \multirow{3}{*}[-1em]{ROS Message Type} & \multicolumn{3}{c}{Encoded Message Size (\emph{B})}\\
      \cmidrule(lr){2-4}
      {} & \textbf{Native ROS} & \textbf{Robofleet} & \textbf{Rosbridge} \\
      {} & Binary & Flatbuffer & JSON \\
      \midrule
      LaserScan & 7,260 & 7,328 & 28,000 \\
      NavSatFix & 127  & 224 & 251 \\
      CompressedImage & 75,188  & 75,224 & 267,650 \\
      \bottomrule
  \end{tabular}
\end{table}

\subsection {Robofleet Client}
Robofleet Client is a lightweight ROS node which is deployed on every robot in the Robofleet system.
It is responsible for facilitating fast, high-throughput communication between the Robofleet Server and the local ROS ecosystem.
This client is implemented in C++, ensuring it has a small compute and memory footprint, as is showcased in our experimental results.
The Robofleet Client, like other components of the Robofleet system, uses WebSocket communication to send encoded ROS messages across the network. It supports a publisher/subscriber model similar to ROS itself, allowing for individual robotic platforms to decide which topics to advertise to the world and which topics from other robots to consume, creating strong topic isolation between robots.
To determine the current connectivity conditions, Robofleet Client uses WebSocket's recommended \textsc{PING}/\textsc{PONG} protocol, which is a form of backpressure monitoring. Robofleet Client sends a \textsc{PING} after each message, and waits for the server's corresponding \textsc{PONG} response, and uses this information to track how many messages are currently in flight. The client has a message scheduler that uses this information to prevent runaway queuing of messages in the case of a network-constrained environment. The following sections describe in detail the configuration, runtime state, and message handling algorithms implemented by the Robofleet Client

\mypara{Client Configuration}
The configuration of Robofleet Client consists of two sets of tuples $\mathbf{T_r}, \mathbf{T_l}$, and a network backpressure threshold $n_T$. The set $\mathbf{T_r} = \{\langle \tau^i, \omega_\tau^i \rangle_{i=1}^{N_r} \}$ specifies $N_r$ remote topics to subscribe to, where $\tau$ is a remote topic name and $\omega_\tau$ is a topic type. The set $\mathbf{T_l} = \{\langle \tau^i, \omega_\tau^i, p_\tau^i, r_\tau^i, d_\tau^i \rangle_{i=1}^{N_l} \}$ specifies $N_l$ local topics to publish to the Robofleet Server. Here $\tau$ is a local topic name, $\omega_\tau$ is a topic type, $p_\tau$ is a non-negative topic priority, $r_\tau$ is a publication rate in Hz, and $d_\tau \in \{ \mathrm{true}, \mathrm{false} \}$ is a flag indicating if $\tau$ should be treated as a \texttt{no\_drop} topic. The threshold $n_T$ determines how many messages may be in-flight between the client and server before the message scheduler begins to limit the rate of message transmission. Each of these values can be tuned on a per-client basis simply by editing a configuration file, to allow robot deployers to optimize the system for their needs.

\mypara{Client State}
During execution, the Robofleet Client's state is comprised of two queues $\mathbf{Q_n}$, $\mathbf{Q_f}$ for message handling, as well as a network backpressure counter $n$ to track network availability. Each queue contains tuples $\langle m_\tau, \tau \rangle$, where $m_\tau$ is a local ROS message on topic $\tau$. The \texttt{no\_drop} queue $\mathbf{Q_n}$ is a first-in-first-out queue containing messages from topics configured with $d_\tau = \texttt{true}$. The \texttt{fair\_priority} queue $\mathbf{Q_f}$ is a priority queue containing one entry per topic configured with $d_\tau = \texttt{false}$. This queue is ordered by $f_\tau = p_\tau \cdot t_\tau$, where $t_\tau$ is the time elapsed since the last message was successfully transmitted on topic $\tau$. When retrieving topics from this queue, they are returned in descending order of $f_\tau$, ensuring that long-waiting and high-priority topics are returned first. The network backpressure counter $n$ is computed as the number of sent messages for which we have not yet received a \textsc{pong} from the server.

\mypara{Publishing Local Messages}
Robofleet Client's message publication flow consists of two primary stages. First, it populates the publication queues by subscribing to the configured local ROS topics and performing rate limiting. Algorithm \ref{alg:localmessage} illustrates the logic for using an incoming ROS message to update local state. It uses the helper function \textsc{PUSH} to add a new element at the appropriate position in the queue, and the helper function \textsc{UPDATE} to perform an in-place update to the message in an existing entry in the queue.

\begin{algorithm}[t]
  \small
  \caption{\textsc{Local ROS Message Handling}}\label{alg:localmessage}
  \begin{algorithmic}[1]
  \State $\textbf{Existing State:}$ Queues $\mathbf{Q_n}, \mathbf{Q_f}$
  \State $\textbf{Existing Configuration:}$ $r_\tau, d_\tau$.
  \State $\textbf{Input:}$ message $m_\tau$, topic $\tau$, elapsed time $t_\tau$
  \If{$t_\tau < \frac{1}{r_\tau}$} \
    \State \textbf{return};
  \EndIf
  \If{$d_\tau == 1$} \
    \State \textsc{Push($\mathbf{Q_n}$, $\langle m_\tau, \tau \rangle$)}
  \ElsIf{$d_\tau == 0$} \
    \If{\textsc{Contains($\mathbf{Q_f}$, $\tau$)}} \
      \State \textsc{Update($\mathbf{Q_f}$, $\tau$, $m_\tau$)}
    \Else \
      \State \textsc{Push($\mathbf{Q_f}$, $\langle m_\tau, \tau \rangle$)}
    \EndIf
  \EndIf
  \end{algorithmic}
\end{algorithm}

The second component of message publication is the message scheduler. By implementing a scheduler in Robofleet
Client, we both enable the prioritization of important messages and prevent
runaway latency caused by queuing in the face of adverse network conditions. Algorithm \ref{alg:localschedule} summarizes how the scheduler uses local state to decide which messages to publish at a given point in time. Here, the helper function \textsc{Publish} is responsible for sending message $m_\tau$ across the WebSocket connection. The helper function \textsc{UpdateBPCounter} is responsible for incrementing $n$ when a new message is sent, and updating it based on the most recently received \textsc{PONG} message. Section \ref{sec:results} presents empirical results showcasing the performance benefits of this algorithm.

\begin{algorithm}[h]
  \small
  \caption{\textsc{Robofleet Client Message Scheduler}}\label{alg:localschedule}
  \begin{algorithmic}[1]
  \State $\textbf{Existing State:}$ Queues $\mathbf{Q_n}, \mathbf{Q_f}$, Integer $n$.
  \If{$n \geq n_T$} \
    \State \textbf{return};
  \EndIf
  \While{$\neg$ \textsc{IsEmpty($\mathbf{Q_n}$)} \textbf{and} $n < n_T$} \
    \State $m_\tau, \tau \gets $ \textsc{POP($\mathbf{Q_n}$)}
    \State \textsc{Publish($m_\tau, \tau$)}
    \State $n \gets \textsc{UpdateBPCounter()}$
  \EndWhile

  \While{$\neg$ \textsc{IsEmpty($\mathbf{Q_f}$)} \textbf{and} $n < n_T$} \
    \State $m_\tau, \tau \gets $ \textsc{POP($\mathbf{Q_f}$)}
    \State \textsc{Publish($m_\tau, \tau$)}
    \State $n \gets \textsc{UpdateBPCounter()}$
  \EndWhile
  \State \textsc{Ping()}
  \end{algorithmic}
\end{algorithm}

\mypara{Receiving Remote Messages}
In order to receive remote messages on a particular topic or set of topics, clients in Robofleet must a send a message of a special type, \texttt{subscription}, to the Robofleet Server. A subscription message $m_S = \langle e_\tau, \mathrm{action} \rangle$ specifies a regular expression $e_\tau$ indicating topic names and an $\mathrm{action} \in \{ \texttt{subscribe}, \texttt{unsubscribe} \}$. Robofleet Client sends a \texttt{subscription} message with $\mathrm{action}=\texttt{subscribe}$ for each  $\langle \tau, \omega_\tau \rangle \in \mathbf{T_r}$ at startup, and corresponding messages with $\mathrm{action}=\texttt{unsubscribe}$ at shutdown. Section \ref{sec:serverhandlemessage} describes how the Robofleet Server handles these messages.

\subsection {Robofleet Webviz}
Robofleet Webviz is a web visualizer built specifically for the task of managing fleets of service-mobile robots capable of long-term autonomy.
To the rest of Robofleet system, it just appears as another client, meaning that Robofleet Webviz is interchangeable with any other client-facing front-end able to communicate with the Robofleet Server. This design decision allows for the Robofleet system to be more easily extended to new domains, as discussed in section \ref{sec:extension}.

\begin{figure}
  \centering
  \begin{subfigure}[b]{\columnwidth}
    \centering
    \includegraphics[width=\textwidth]{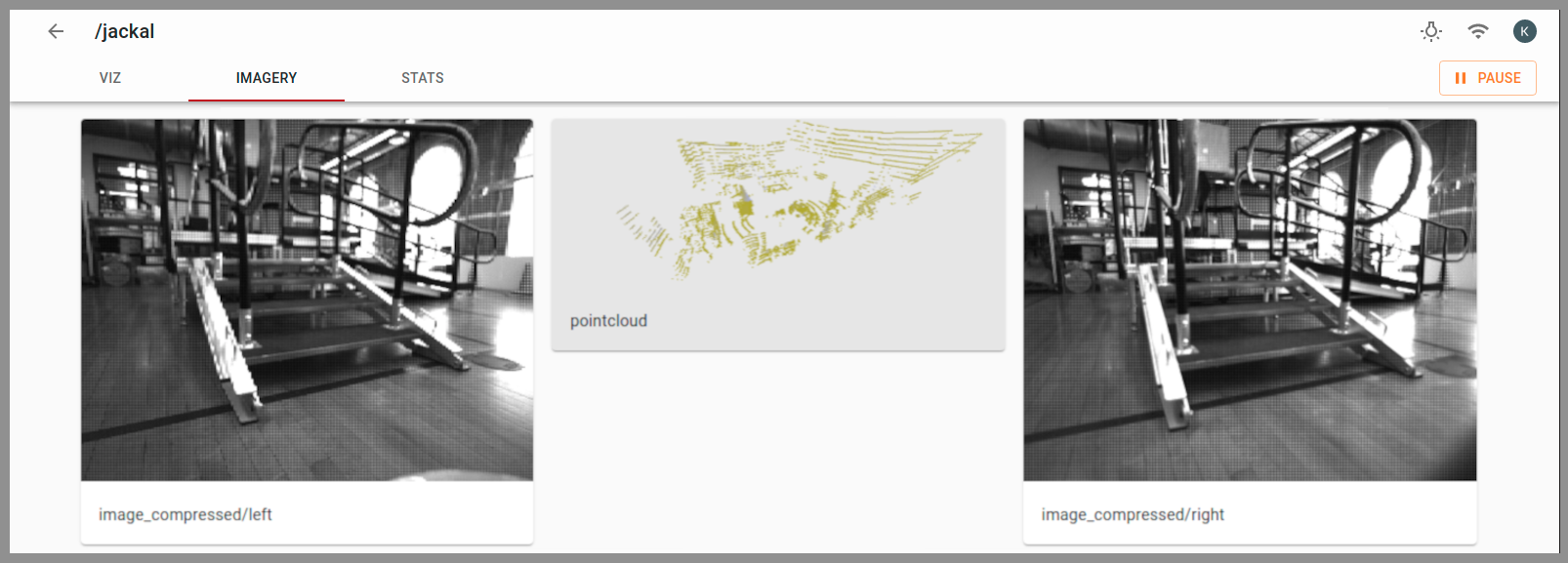}
  \end{subfigure}
  \par\medskip
  \begin{subfigure}[b]{\columnwidth}
    \centering
    \includegraphics[width=\textwidth]{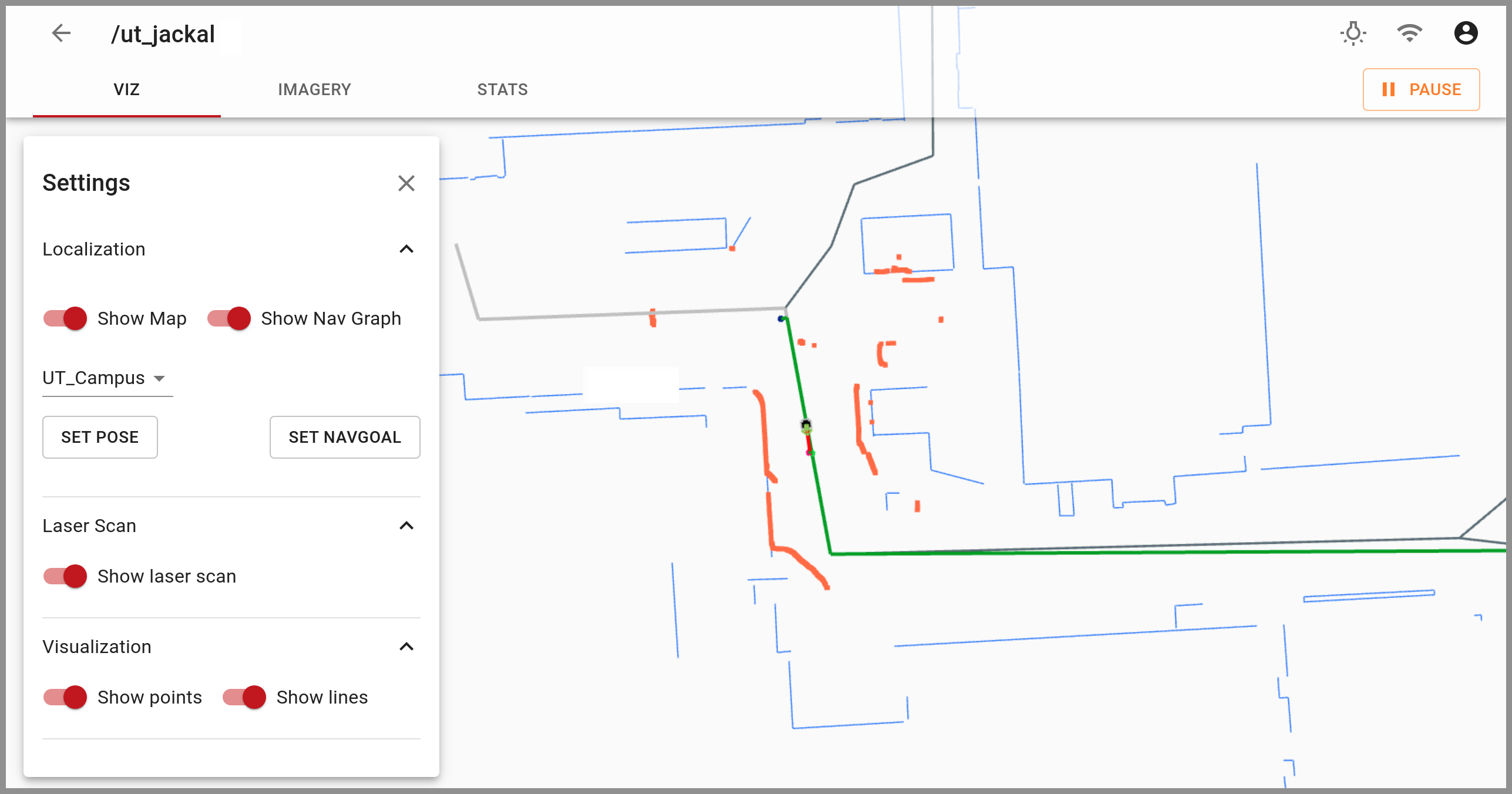}
  \end{subfigure}
  \caption{Robofleet Webviz Interface. (Top) Stereo camera images and 3d lidar point cloud. (Bottom) Visualization showing 2d lidar scan (orange), localization estimate, and navigation plan (green).}
  \label{fig:webviz}
\end{figure}

Robofleet Webviz has no dependence on the ROS ecosystem, allowing for it to be deployed on any browser-enabled device, which is ideal for remote monitoring and tasking. This web interface contains an interactive list of currently live robots, as well as coarse status information from recently-deployed robots.
For each live robot, authorized users can use Webviz to view
\begin{inparaenum}[a)]
  \item camera images
  \item 2d lidar
  \item 3d lidar
  \item odometry information
  \item localization estimates
  and \item arbitrary visualizations similar to ROS's \texttt{Marker} messages 
\end{inparaenum}. Finally, Webviz has the ability to send navigation and localization commands to the robot. Figure \ref{fig:webviz} shows sample screenshots from the Webviz interface.

\subsection {Robofleet Server}
The Robofleet Server runs on a single, networked server and is the central point of communication for the Robofleet system.
It is responsible for authenticating and relaying WebSocket messages from robots, managing subscription information, maintaining persistent robot information, and authenticating and servicing user requests.
Robofleet Server uses static IP address based authentication for robot clients. To ensure the security of the system, we deploy the robots on a VPN with crypto-based routing, WireGuard \cite{Donenfeld2017WireGuardNG}. This VPN uses private/public key pairs to validate the identity of machines on the network, preventing the spoofing of IP addresses and allows for simple, low-compute authentication of robots.
For incoming requests from the open internet, for example from Robofleet Webviz, Robofleet supports Google SSO authentication. The following sections describe in detail the configuration, runtime state, and message handling algorithm of the Robofleet Server.

\mypara{Server Configuration}
Robofleet Server configuration primarily involves defining an authorization map $\mathbf{M} : u \rightarrow \{\langle e_\tau^i, \mathrm{op}^i \rangle\}_{i=1}^{N_a}$ from sender identity $u$ to a list of $N_a$ authorization rules. Sender identity $u$ is either an IP address, a range of IP addresses, or an email address. Authorization rules are tuples where $e_\tau$ is a regular expression to specify the topics to which this rule should apply, and $op \in \texttt{\{send, receive, both\}}$ specifies the permissions associated with this rule.

\mypara{Server State}
While running, the Robofleet Server retains a subscription map $\mathbf{S}$ from each topic $\tau$ to a list of the of all active clients subscribing to $\tau$. The server's message handling algorithm is responsible for updating $\mathbf{S}$ according to incoming subscription messages.

\mypara{Message Handling Algorithm}
\label{sec:serverhandlemessage}
Algorithm \ref{alg:serverhandlemessage} describes how the Robofleet Server handles new incoming messages from clients. The helper function \textsc{IsAuthorized} first extracts relevant sender information from the request, then uses the authorization map $\mathbf{M}$ in conjunction with the requested topic $\tau$ and operation $\mathrm{op}$ to decide if the given action should be permitted. If the authorization check succeeds, Robofleet Server uses the helper function \textsc{IsSubscriptionMessage} to decide whether to update its subscription map $\mathbf{\mathbf{S}}$ (lines 7-12), or relay the message to the appropriate clients in $\mathbf{S}[\tau]$ (lines 14-18).

\algnewcommand\algorithmicforeach{\textbf{for each}}
\algdef{S}[FOR]{ForEach}[1]{\algorithmicforeach\ #1\ \algorithmicdo}

\begin{algorithm}[h]
  \small
  \caption{\textsc{Robofleet Server Message Handling}}\label{alg:serverhandlemessage}
  \begin{algorithmic}[1]
  \State $\textbf{Existing State:}$ Subscriptions map $\mathbf{S}$
  \State $\textbf{Input:}$ message $m_\tau$, topic $\tau$, sender $u$
  \If{!\textsc{IsAuthorized($u$, $\tau$,  \texttt{send})}} \
    \State exit
  \EndIf
  
  \If{\textsc{IsSubscriptionMessage($m_\tau$)}} \
    \State $\tau_s, \mathrm{action} \gets $ \textsc{ExtractSubscriptionInfo($m_\tau$)}
    \If{$\mathrm{action} = \texttt{subscribe}$ \textbf{and}\\
         $\qquad \qquad$ \textsc{IsAuthorized($u$, $\tau_s$, \texttt{receive})}} \
      \State $\mathbf{S}[{\tau_s}] \gets \mathbf{S}[{\tau_s}] \cup \{u\}$
    \ElsIf{$\mathrm{action} = \texttt{unsubscribe}$} \
      \State $\mathbf{S}[{\tau_s}] \gets \mathbf{S}[{\tau_s}] \setminus \{u\}$
    \EndIf
  \Else \
    \ForEach {$u_s \in \mathbf{S}[\tau]$}
      \If{\textsc{IsAuthorized($u_s$, $\tau$, \texttt{receive})}} \
        \State \textsc{Send($m_\tau$, $u_s$)}
      \EndIf
    \EndFor
  \EndIf

  \end{algorithmic}
\end{algorithm}

Robofleet's single-server architecture provides a few key benefits to the system. First, it inherently provides message deduplication in a many-robot scenario.
In a traditional ROS-based approach, each robot is responsible for sending its data stream to all its consumers, which can be costly for network-constrained mobile platforms. In Robofleet, each robot only sends a single data stream, to the server, which then relays this data to the relevant consumers. Section \ref{sec:multirobot} presents empirical evidence of the benefits of this architecture for multi-robot systems.
Additionally, as the only entry point into the system, authentication can be implemented only at the server level, leaving clients free to interact without having to dedicate compute resources to security concerns.

%% file: results.tex
\section{Experimental Results}
\label{sec:results}

To validate the performance of the Robofleet system, we performed a series of benchmarks comparing its performance to the two most widely-used robot communication architectures today: ROS Native and Rosbridge. We first show that Robofleet uses only a small fraction of the onboard compute resources of Rosbridge. We then compare Robofleet to both existing systems by measuring throughput (messages transmitted in a fixed time frame) and latency (time for a message to reach its destination) under ideal network conditions. We continue these comparisons for simulated adverse network conditions, as well as real-world robot deployments and multi-robot scenarios.

\subsection{Computational Load}

\begin{table}[h]
  \caption{\textsc{Compute Usage over 15 seconds of 10 Hz \texttt{PointCloud2} message transmission}}
  \label{tab:compute_usage}
  \centering
  \resizebox{\columnwidth}{!}{
    \begin{tabular}{lrrrr}
        \toprule
        \multirow{2}{*}[-1em]{\makecell[l]{Comm. \\ Client}} & \multicolumn{2}{c}{\makecell{\textbf{Desktop Workstation} \\ Intel i7-9750H CPU}} & \multicolumn{2}{c}{\makecell{\textbf{Mobile Robotic Platform} \\ Jetson TX2}} \\
        \cmidrule(lr){2-3}
        \cmidrule(lr){4-5}
        {} & CPU Usage (\%)  & Memory (\texttt{MB}) & CPU Usage (\%) & Memory (\texttt{MB}) \\
        \midrule
        \textbf{Robofleet} & $\mathbf{1.3 \pm 0.0}$ & $\mathbf{452.5 \pm ~~~0.0}$ & $\mathbf{15.1 \pm 0.5}$ &  $\mathbf{425.6 \pm 0.0}$ \\
        Rosbridge & $16.6 \pm 4.1$  & $1,339.6 \pm 101.9$ & $155.9 \pm 0.0$ &  $732.9 \pm 0.0$ \\
        \bottomrule
    \end{tabular}
  }
\end{table}

Robotic platforms are compute-bound due to the abundance of concurrent tasks they must perform, such as perception and navigation, making it critical that processes running on robots are as efficient as possible. This experiment compares the computational load of running the Robofleet client with running a sample Rosbridge client (in this case, \texttt{RosLibPy}).
Note that both of these clients run in addition to native ROS, which is not included in this comparison. 

This experiment was conducted by playing a 15 second sequence of \texttt{PointCloud2} messages from a fixed bag file at 10 Hz, while running a communication client responsible for sending these messages to a remote machine. We measure the CPU utilization and virtual memory footprint on a desktop workstation and a mobile robotic platform, repeating the experiment 10 times to collect information on variance. Table \ref{tab:compute_usage} shows the aggregated results of this experiment, demonstrating Robofleet's significantly smaller computational load when compared to Rosbridge. In this table CPU usage is computed as $\frac{\mathrm{CPU}\:\mathrm{Time}}{\mathrm{Wall}\:\mathrm{Clock}\:\mathrm{Time}}$, allowing usage to exceed $100\%$ on multi-core processors.

\subsection{Message Transmission}

The remainder of the experiments conducted compare the transmission properties of Robofleet with existing systems. Throughout these experiments we use a sample set of ROS messages of standard types including \texttt{PointCloud2} and \texttt{Odometry}, and the Robofleet system is configured with $n_T = 1$. We have observed behavior similar to what is presented here with a variety of topic types, though the magnitude of the differences between the systems changes. We measure latency as the time it takes for a message sent from one client to reach another, and throughput as the number of messages that reach a client per second.

\begin{figure}
  \centering
  \includegraphics[width=\columnwidth]{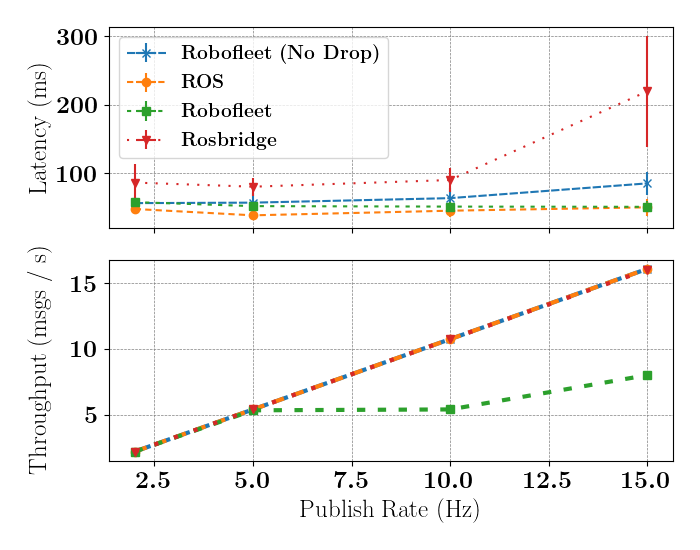}
  \vspace{-1.75em}
  \caption{Latency and Throughput for \texttt{PointCloud2} Messages over 100Mbit client network.}
  \label{fig:latency_throughput_simple}
\end{figure}

Figure \ref{fig:latency_throughput_simple} illustrates the effect of increasing
the publication rate of point cloud messages given a stable 100Mbit connection
for each client. The system labeled \textbf{Robofleet (No Drop)} is a Robofleet
system where all topics are configured with the \texttt{no\_drop} flag set to
\texttt{true}, meaning no messages will be dropped to alleviate network
backpressure. These results clearly show the explicit trade-off being made by
the Robofleet system, sacrificing message throughput to ensure low latency and
message liveness. For these large messages under ideal network conditions,
Native ROS's binary encoding allows it to outperform both Robofleet and
Rosbridge.

\mypara{Adverse Network Conditions} When navigating in the wild, robots will often experience a variety of network adversities, such as temporary network dropout and gradual network degradation.  We performed simulated experiments to demonstrate how Robofleet compares to existing systems in its handling of these conditions. To perform these experiments, we leveraged Linux's \texttt{tc} utilities to limit network traffic in a controlled manner on the client sending the messages. As before, these experiments measure transmission from a single robot client to a remote machine.

\begin{figure}
  \centering
  \includegraphics[width=\columnwidth]{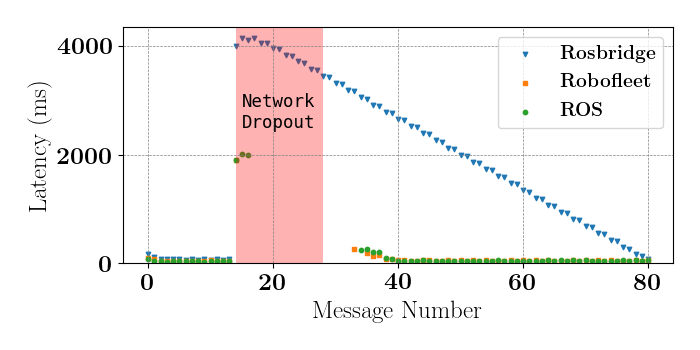}
  \vspace{-2em}
  \caption{Latency under simulated network dropout}
  \label{fig:latencydropout}
\end{figure}

Figure \ref{fig:latencydropout} presents a sample time series of message transmission in the presence of a temporary network dropout, simulating a mobile robot switching access points, or briefly entering a Wi-Fi dead zone. Because Rosbridge has no client-side traffic control, the WebSocket connection on the client's side quickly becomes backlogged with messages. By contrast, Robofleet simply drops messages that will not reach the destination in a reasonable time, allowing it to quickly recover to reasonable message latencies, retaining performance similar to native ROS. Critical messages are retained on the queue to be delivered once the network returns to normal. We observe that after the network dropout, Robofleet is able to return to nominal latency around 5 times more quickly than Rosbridge.

\begin{figure}
\centering
\includegraphics[width=\columnwidth]{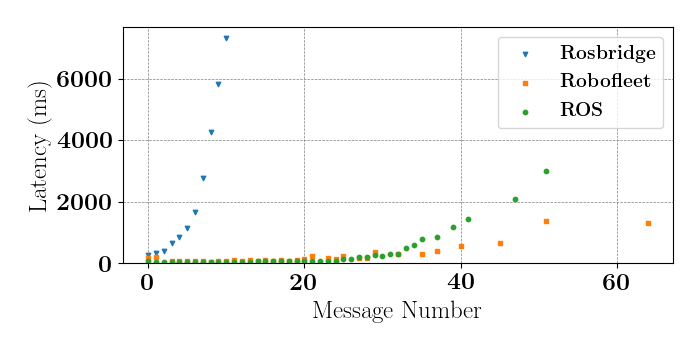}
\vspace{-2em}
\caption{Latency under simulated bandwidth degradation}
\label{fig:latencydegradation}
\end{figure}

Figure \ref{fig:latencydegradation} shows a similar time series demonstrating the communication behavior under bandwidth degradation, where we linearly decrease the available bandwidth on the sending client from 100 Mbit down to 4Mbit over the course of the 15 second test, simulating a robot traveling increasingly far from its nearest access point, and therefore observing degrading network quality. We see in this case that due to the Rosbridge's large encoded message size, it quickly experiences major performance degradation in the presence of this constrained network. We additionally observe that ROS retains acceptable performance longer due to its more compact message size, but eventually also degrades due to a lack of client-side backpressure monitoring.

\begin{figure}
  \centering
  \includegraphics[width=\columnwidth]{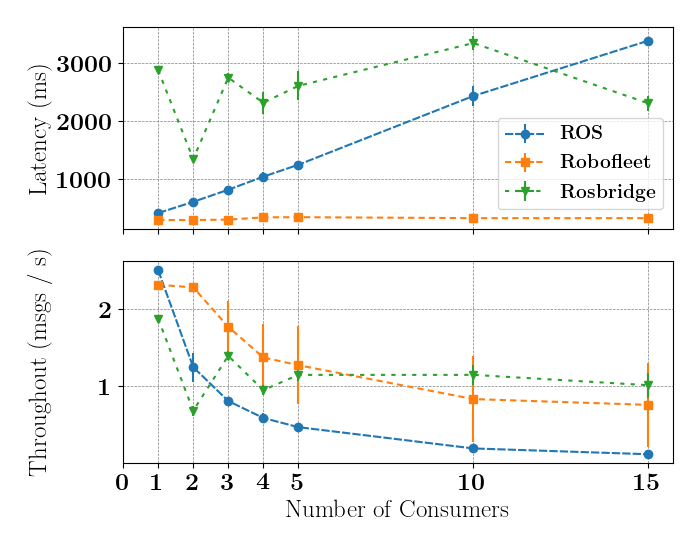}
  \vspace{-2em}
  \caption{Latency and Throughput \textit{vs.} Number of Consumers for 3D point clouds at 5 Hz over 12Mbit client network.}
  \label{fig:latency_throughput_multirobot}
\end{figure}

\mypara{Multi-Robot Experiments} \label{sec:multirobot} To demonstrate the challenges a native ROS-based system encounters in the presence of multiple robots, we performed a simple experiment where a single robot client sent point cloud messages at 5 Hz to a varying number of clients. In this situation, the sending robot client was allowed a 12 Mbit network connection, to simulate upload the upload speed of a typical 4G LTE connection. Figure \ref{fig:latency_throughput_multirobot} illustrates the message transmission latency and throughput as the number of consumers increases. Due to the peer-to-peer nature of ROS native, for each new consumer, the robot client needed to send a new data stream across the wire. This results in the duplication of data, causing the linear increase in latency observed in the experiment. By contrast, in Robofleet and Rosbridge, there is only a single stream of data sent from the robot to a central server, which then relays the information to any relevant consumers on a significantly less constrained network. It is worth noting that for the large \texttt{PointCloud2} messages in this test, Rosbridge performs significantly slower than ROS Native in the single robot scenario, but becomes comparable as the number of robots increases. In addition, Robofleet dramatically outperforms both systems in terms of latency, while retaining comparable throughput, as the number of consumers increases.

\begin{figure}
  \centering
  \includegraphics[width=\columnwidth]{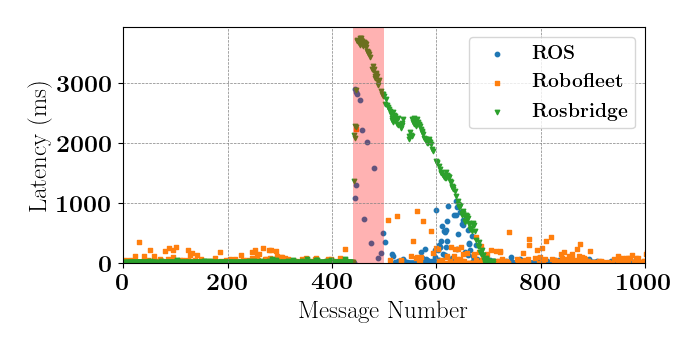}
  \vspace{-2em}
\caption{Message latency during on-campus deployment with localization, image,
and laser scan streams.}
  \label{fig:livelatency}
  \end{figure}

 \mypara{Real-World Deployment} To validate the results of our synthetic experiments, we deployed Robofleet, Rosbridge, and native ROS on a robot deployed at a university campus. For this experiment, we deployed each system on a Clearpath Husky connected to the campus Wi-Fi network, and repeated a short 50m trip, transmitting its estimated localization information as \texttt{PoseStamped} at 10 Hz, the current lidar scan as \texttt{LaserScan} at 10 Hz, and images from one of its cameras as \texttt{CompressedImage} at 2 Hz. Figure \ref{fig:livelatency} shows the observed latency of messages sent during this deployment. Along the trajectory, there was one primary region (marked in red), which suffered from poor Wi-Fi coverage, and we observed very large spikes in network latency when the robot entered this area across all systems, starting around message number 430. However, we clearly see that these negative conditions affected Rosbridge for a significantly longer period of time than the other two systems, as expected from our synthetic experiments.

In summary, the experimental results presented here show that Robofleet outperforms Rosbridge for message transmission, and has performance comparable to that of native ROS in most conditions. Its performance relative to these two systems increases dramatically when there are network constraints such as limited bandwidth or intermittent connectivity issues. Additionally, Robofleet dramatically outperforms native ROS in multi-robot scenarios.

%% file: future_work.tex
\section{Discussion \& Future Work}

The development and deployment of Robofleet on a fleet of autonomous robots has uncovered key points of discussion and areas of future work. First, we discuss the potential to extend Robofleet to domains in robotics beyond long-term autonomy. Then, we explore the similarities between Robofleet communication and the ROS2 transport layer, and discuss their symbiotic nature.

\subsection{Extension to New Domains}
\label{sec:extension}

The communication protocol underlying Robofleet is domain agnostic. By design, there is nothing in the Robofleet's communication architecture which limits its use to communication for autonomous robot platforms, and supporting new domains would simply involve extending the encoding library to support any new topic types. This process involves auto-generating type definitions from the ROS message definitions, and implementing $\texttt{encode}$ and $\texttt{decode}$ functions for these messages, and is extensively documented in the Robofleet code repository. An additional possible extension is implementing a remote procedure call mechanism similar to ROS Actions to support use cases for which the publisher/subscriber model is insufficient.

In addition to the communication protocol, Robofleet provides the Robofleet Webviz interface, which is tailored towards autonomy tasks. However, alternative interfaces can be implemented to support new use cases, with the only dependency being the Robofleet encoding library, which supports both TypeScript and C++.
Alternate front-ends to the system are being actively developed, including a
native Microsoft
HoloLens~\footnote{\url{https://www.microsoft.com/en-us/hololens}} AR interface
for robotic communication (Video: \url{https://www.youtube.com/watch?v=F36ZeBU7R6A}).

\subsection{ROS2 Ecosystem}
The design of ROS2's transport layer recognized many of the shortcomings of traditional ROS, and the design of the new transport layer shares conceptual similarities with Robofleet. It introduces the concept of ``Quality of Service" \cite{ROS2QOS}, which coincides with the  message transmission concepts introduced in this paper. As such, a future adaptation of Robofleet could fit into this new ecosystem in a way coherent with ROS2's design principles.

%% file: conclusion.tex
\section{Conclusion}

In this paper we have highlighted the need for secure, low-latency, reliable two-way robotic communication. We then presented Robofleet, an open-source communication and management system for fleets of autonomous mobile robots. Our experimental results highlighted the low computational overhead of the Robofleet system, demonstrated its resilience to a variety of adverse network conditions, and showcased its ability to scale to large numbers of robotic clients. Finally, we discussed future extensions for Robofleet, which will bring its benefits to the wider robotics community.

%% file: acknowledgements.tex
\section{Acknowledgements}
This work has taken place in the Autonomous Mobile Robotics Laboratory (AMRL) at UT Austin.  AMRL research is supported in part by NSF (CAREER-2046955,
IIS-1954778, SHF-2006404), ARO (W911NF-19-2-0333), DARPA (HR001120C0031),
Amazon, JP Morgan, and Northrop Grumman Mission Systems.
The views and conclusions contained in this document are those of the authors alone. The authors would like to thank Jack Borer, Max Svetlik, Can Pehlivanturk, and UT's Nuclear and Applied Robotics Group for helping validate Robofleet by deploying it on multiple robotic platforms.